%% file: conference_101719.tex
\documentclass[conference]{IEEEtran}
\IEEEoverridecommandlockouts
\usepackage{biblatex}
\usepackage{tabularx}
\usepackage{multirow}
\usepackage{amsmath,amssymb,amsfonts}
\usepackage{algorithmic}
\usepackage{graphicx}
\usepackage{textcomp}
\usepackage{xcolor}
\def\BibTeX{{\rm B\kern-.05em{\sc i\kern-.025em b}\kern-.08em
    T\kern-.1667em\lower.7ex\hbox{E}\kern-.125emX}}
\begin{document}
\title{Explainable Deepfake Video Detection using \\ Convolutional Neural Network and CapsuleNet\\
}
\author{\IEEEauthorblockN{Gazi Hasin Ishrak\IEEEauthorrefmark{1},
Zalish Mahmud\IEEEauthorrefmark{1}, MD. Zami Al Zunaed Farabe\IEEEauthorrefmark{1}, Tahera Khanom Tinni\IEEEauthorrefmark{1},\\Tanzim Reza \IEEEauthorrefmark{1} and Mohammad Zavid Parvez \IEEEauthorrefmark{1}}
\IEEEauthorblockA{\IEEEauthorrefmark{1}Department of Computer Science and Engineering, School of Data and Sciences, BRAC University, Bangladesh\\
Email: 
\IEEEauthorrefmark{1}gazi.hasin.ishrak@g.bracu.ac.bd,
\IEEEauthorrefmark{1}zalish.mahmud@g.bracu.ac.bd,
\IEEEauthorrefmark{1}md.zami.al.zunaed.farabe@g.bracu.ac.bd,\\
\IEEEauthorrefmark{1}tahera.khanom.tinni@g.bracu.ac.bd,\\
\IEEEauthorrefmark{1}tanzim.reza@bracu.ac.bd,
\IEEEauthorrefmark{1}zavid.parvez@bracu.ac.bd}}
\maketitle

\begin{abstract}
\input{chapters/Abstract}

\end{abstract}

\begin{IEEEkeywords}
Deepfake, CapsuleNet, GAN, Computer-vision, LSTM, CNN, XAi, Explainable-Ai,Grad-Cam
\end{IEEEkeywords}

\section{Introduction}
\input{chapters/introduction}

\section{Literature Review}
\input{chapters/Literature_Review}

\section{Methodology}
\input{chapters/Work_Plan}

\section{Dataset}
\input{chapters/Dataset}
\section{Architecture}
\input{chapters/architecture}
\input{chapters/methodology}
\section{Implementation \& Results}
\input{chapters/Implementation}

\section{Comparative Analysis}
\input{chapters/Analysis}

\input{chapters/conclusion}

\section*{Acknowledgment}
\input{chapters/Ackowledgement}

\printbibliography

\end{document}

%% file: chapters/Abstract.tex
The term ‘Deepfake’ comes from the deep learning technology. Deepfake technology can easily and smoothly stitch anyone into any digital media where they never took part in reality. The key components of deepfake are machine learning and Artificial Intelligence (AI). At the beginning deepfake was introduced for research, industrial uses and entertainment purposes. The capabilities of deepfakes have existed for decades but the creations were not as realistic as they are today. As time passes, deepfakes are also improving and creating such things which are hard to identify as ‘real’ or as ‘fake’ with bare eyes. Furthermore, the new technologies now allow anyone to make deepfakes even if the creator is unskilled. The ease of accessibility and the increase of availability of deepfake creations have raised the issue of security.The most highly used algorithm to make these deepfake videos is GAN (generative Adversarial network), which is basically a machine learning algorithm which creates a fake image and discriminates itself to reproduce the best possible Fake frame or video. Our Primary goal is to use CNN (Convolutional Neural Network) and CapsuleNet with LSTM to distinguish which frame of the video was generated by the deepfake algorithm and which was the original one. We also want to find out why our model predicted the output of detection and analyze the patterns using Explainable AI. We want to apply this approach to develop a transparent relation between AI and human agents and also to set an applicable example of explainable Ai in real-life-scenarios.


%% file: chapters/Introduction.tex
With the advancement of computer graphics, computer vision and machine learning, we now have the ability to create ultra realistic media that replicates the real world environment. Not to mention the digital game industry and cinematographic industry are thriving to get in hand with the latest computer graphic tools to generate ultra-realistic graphics. This competition for perfect realistic image/video quality results in more and more powerful and precise computer graphics (CG) methods for digital scenario generation. With the help of high processing power in current computers, such methods can create results so monumental that the viewer may find it hard to recognize them. However, as pointed out by Holmes et.al.\cite{holmes2016assessing}, once the perfect CG image generation goal is achieved, it brings a challenge with itself in the field of technology, to identify or to discriminate between a photo-generated (PG) and a CG method 
Deepfake is an image/video manipulation technique that uses different algorithms and image processing techniques. It is made by switching a person's face in a video with another person’s face. Deep Fakes are becoming more popular and easier to execute as video manipulation becomes more automated. Recently, these manipulated images/videos are being used to create political distress, blackmail someone, or gather fake terrorism events.
Human beings are at risk to be the victims of deepfake if deepfakes are left undetected. Unfortunately, even if the content is detected as fake meanwhile many damages would have been already occurred in millions of eyes. Ergo, deepfakes are increasing the public discomfort and distrust in all spheres. Celebrities are being the victims of pornography, politicians are being the victim of defamation in public, creating conflicts in politics. Not to mention business negotiations are falling because of rivalry since videos of executives are getting viral speaking bad things about executives of other companies. Ultimately, no one is safe from deepfake as it is being used for revenge porn, cyberbullying, blackmailing, etc.
\section{Research Problem}
Technology has advanced to the point where generating fake media has become very easy. Deepfake technologies can now make things happen which never even occurred in reality. Some well known deepfake techniques are face swapping \cite{korshunova2017fast}, face re-enactment \cite{thies2016face2face}, voice synthesis \cite{lorenzo2018voice} and so on. At the beginning, deepfakes appeared for the use of good purpose. It has a great impact on film industry where deepfake is used for dubbing or de-aging of actor/actresses and so on. However, deepfake also bring some risk factors as today it is also used for many ill will. For example, In 2017 face swapping of actors in pornography was used which defamed many celebrities. According to the Deeptrace report, 96\% of deepfake videos online were pornographic. Unfortunately, the malicious work in the dark web was not the only domain. Gradually, it expanded into the political sphere. For example, a video got viral of the former president Barack Obama where an American actor named Jordan  Peele  acted as the voice of the president and then later on it was synthesized over the original video of President Barack Obama \cite{obama}. The goal of that video was to create awareness of the advancements in deepfake technology. Another recent example is, during COVID-19 outbreak the Belgium’s Prime Minister was deepfaked in a clip and it made him say that instant action is imperative to tackle the climate change and environmental crisis. The deepfake video was released by a political group of belgium \cite{galindo_2020}. Sadly, this is not the only case in the political sphere, some other cases exist where politicians used the fact of deepfake to discredit real video evidence. Moreover, people around the world are taking advantage of deepfake to defame others or to trick people. With the technological advancement deepfake is learning to make more realistic things as time passes. That is why it has become very difficult to discern between what is ‘real’ and what is ‘fake’. To overcome these serious and dangerous issues we require deepfake detection methods which can recognize subtle hints to distinguish between ‘real’ and ‘fake’. Therefore, many deepfake detection techniques have been introduced so far. For instance, detecting eye blinking patterns, face manipulation, headpose, audio-spoof, face-warping and so on. Researchers have introduced many approaches to detect deepfakes which uses Convolutional Neural Networks (CNN). Nevertheless, CNN lack in detecting fake videos when they are applied to inverse graphics. After every convolution layer the pooling layers of CNN cause loss of information and they are unable to identify a face if the features are not in correct order. CNN face difficulties while classifying tilted or rotated images. Besides, beacuse of the maxpool layer CNN becomes considerably slower. In order to overcome the shortcomings we have proposed a hybrid model in our paper.
 
\section{Research objectives}
Our objective of this research is to develop a model that will not just accurately detect deepfake videos but also explain why the video was detected to be a fake video. This will allow us to learn which facial landmarks are most commonly used to create the deepfake video. Moreover, We aim to create a model with such optimization and adaption methods that it can recognize deepfake videos even if the video quality is low, the subject is moving, or the actual video frame is rotated. The work can be extended by studying the patterns that will be generated by our proposed model to identify the primary approaches for making deepfake videos using the methods of face swapping and face warping technology. With the help of this data, we will be able to expose the basic facial landmarks being targeted in making of deepfake videos. This can help us to create new methods or techniques that will be targeting the detected landmarks to identify the authenticity of a video. 
The primary goal of our research are:

\begin{itemize}
 \item  Understanding the current methods that applied to make deepfake videos and how to detect the deepfake video.
\item Propose a model that will detect deepfake videos for the current available datasets.
\item Optimize the model to detect deepfakes in different video properties.
\item Implement explainable Ai to justify the models classification outcomes.
\end{itemize}


%% file: chapters/Literature_Review.tex
\subsection{Deepfake} 
For the last couple of years the number of Fake videos and images are not only increasing but they are also getting more realistic than ever before. The main Technology behind this is Deepfake technology. This technology uses Deep learning models like autoencoders and Generative Adversarial Network (GAN) to detect facial features from an image or a video frame and replace that with a target person’s face. This Deep learning approach makes the creation of fake videos more easier than traditional editing images or videos via software, which requires a lot of skills and a huge amount of time to create realistic fake images or videos. To train these Deep Fake models requires huge amounts of images or video data. Now a days most public figures images and videos are publicly available on the internet. So, These days using existing deepfake tools anyone can create fake videos of public figure or influencer which can be used for spreading propaganda and false news to mislead people. 

\subsection{Deepfake techniques}
Among a number of ways of generating Deef-Fake videos, here first we will know about the traditional ones and then the most recent Adversarial methods. We will have a quick overview of how these models function and their algorithms for detecting fake videos.

\subsubsection{Face swap}
The face of a target person is layered on the face of a source in this graphical approach. Post processing the overlapping faces such as, color correction, blending edges, matching the source’s facial contour etc. Gives the final output of a face swapped deep fake video. This approach is implemented in various deepfake generation methods such as FaceSwap, DeepFaceLab, DFaker.

\subsubsection{Face re-enactment}
Methods of facial re-enactment cause the target to act and speak like the source. For example, consider the video in which former US President Barack Obama was shown to speak whatever the source video said \cite{obama}. These approaches recognize facial critical points and modify the target's facial points to match the source's facial movements by modeling both the source and target faces. Face2Face \cite{thies2016face2face}, a real-time facial reenactment system, is a common approach in which the facial emotions of a source face are transferred onto the target's face while the target's facial characteristics are preserved.

\subsubsection{Adversarial methods}
Deepfake methods based on Generative Adversarial Networks (GANs) are very recent. GANs are made up of two competing neural networks: a generator G that generates fake samples that resemble real samples from a target dataset, and a discriminator D that tries to identify fakes from real samples. In order to improve the generator and the discriminator networks, both of them are trained simultaneously. The generator can then be used to build realistic-looking examples after it has successfully converged.

\subsubsection{AI-based Video Synthesis Algorithm}
The latest breakthrough in deep learning models, particularly generative adversarial networks (GANs) \cite{goodfellow2014generative}, have emerged as a new generation of AI-based video synthesis techniques. The GAN model inspired many following works for image synthesis, such as \cite{denton2015deep} \cite{shrivastava2017learning}. Liu et al \cite{liu2017unsupervised} proposed an unsupervised image to image translation framework based on coupled GANs. The goal of this algorithm is to learn integrated representation of images in different domains. This algorithm is the premise of deepfake algorithm. Zhu et al. \cite{zhu2017unpaired} proposed Cycle-GAN which cycles through consistent loss to increase the performance of GAN. Bansal et al. \cite{bansal2018recycle} modified the model to Recycle-GAN by implementing temporal information and spatial cues with conditional GAN. StarGan \cite{choi2018stargan} is an approach where it learns the depiction across multiple domains by using only a single generator and discriminatior.

\subsection{Deepfake detection}
\subsubsection{Mesonet}
MesoNet \cite{afchar2018mesonet} is a CNN-based model containing limited number of layers that pays attention on the mesoscopic features of image Since human eyes fail to discern fake images at the macroscopic level and image noise based microscopic analysis is also not possible due to video compression, they use an interim method to focus on mesoscopic analysis. Meso-4, which employs traditional convolutional layers, and MesoInception-4, which is built on Inception modules \cite{szegedy2017inception}, are the two models available. These models attained an average detection rate of 98\% after being trained on fake videos from the internet.

\subsubsection{XceptionNet}
XceptionNet \cite{chollet2017xception} one of the most promising deep neural network architecture for feature extraction. It can be used in traditional studies as well as in submerged space. This architecture of Xception net not only has skip connection like ResNet \cite{he2016deep} but also has an convolutional structure like inception net \cite{he2016deep} It contains a modified version of convolutional layer which is depth-wise separable to improve the performance by reducing the parameters. 

\subsubsection{Convolutional LSTM}
Convolutional LSTM \cite{guera2018deepfake} is a temporal-aware pipeline for detecting and validating deepfakes built from videos collected from several video-hosting services.The suggested architecture combines convolutional layer and LSTM layer where the convolutional layer extracts features from frames and LSTM is used for sequence analysis of those features. In the making of deepfakes each frame generates a new face, which will have some inconsistencies compared to the previous frames and eventually will lack temporal awareness among the frames. Fake videos are detected by analysing the temporal inconsistencies between the sequences of frames such as  artificial illumination in different sections, flickering.Their dataset achieved 97\% accuracy by accumulating convolutional LSTM.

\subsubsection{RecurrentConvolutional}
RecurrentConvolutional \cite{sabir2019recurrent} exploits the  temporal inconsistencies across several frames induced by frame-by-frame changes. DenseNet and ResNet are the CNN models used, with DenseNet outperforming ResNet.
Convolutional LSTM \cite{guera2018deepfake} uses pre-trained CNNs, whereas Recurrent Convolutional models are trained end-to-end. On the FF++ data-set, DenseNet with alignment and bi-directional recurrent network gives the best performance with accuracy of 96.9\%.

\subsubsection{Resampling Detection}
The deepfake video generation process introduces artifacts due to affine transformations to the modified face.  Detecting transforms or the resampling algorithms has been meticulously studied \cite{prasad2006resampling} \cite{kirchner2008hiding}.These methods target to approximate the resampling operation from the whole image. After that, the approach was used to compare the regions of possible synthesized faces. As the latter is expected to be free of such artifacts, the existence of such artifacts in the former is referred to as a modified video.

\subsubsection{FakeSpotter}
Wang et al. \cite{wang2019fakespotter} proposed a new approach by inspecting neuron behaviors of a committed CNN to detect faces which are generated by deepfake technologies. The comparison with Zhang et al \cite{zhang2019detecting} resulted in an average detection accuracy of more than 90\%.

\subsubsection{Convolutional Trace}
GANs are used to generate deepfakes. Once trained, the key element involved in the image generation  is the generator G which is composed of Transpose Convolutional Layers \cite{radford2015unsupervised}.
Kernels are applied to the input images keeping the similarity to kernels in Convolutional Layer. However, they act inversely in order to achieve and output larger but proportional to the input dimensions. Therefore, the picture production pipeline differs from the pipeline frequently employed in camera devices, in which each stage contributes common noise, which is subsequently utilized for naive image fraud detection \cite{zhang2019detecting}. Nonetheless, the picture production process, which employs GAN Transpose Convolutional layers, should be consistent and observable in local correlations of pixels in the spatial RGB space. To locate these traces, the Expectation Maximization (EM) algorithm \cite{moon1996expectation} is used to construct a conceptual mathematical model capable of capturing pixel correlation in spatial pictures and distinguishing between two distributions: the genuine and the deepfake. The EM result is a feature vector that depicts the structure of the Transpose Convolutional Layers that were used throughout the generation process of the deepfake image.

\subsection{Related works} 
In \cite{agarwal2020detecting}, Agarwal and Farid proposed a method that exploits the fact that the dynamics of the mouth shape (visemes) are sometimes inconsistent with a spoken phoneme. They mainly focused on the visemes that associate with the words having sound M(mama), B(baba) and P(papa), which are the only phonemes that require the mouth to be completely closed in order to pronounce. This described forensic technique is to detect lip-sync deep fakes, categorized as high-level detection techniques, which consists of inconsistencies in eye blinks \cite{li2018ictu}, head-pose \cite{yang2019exposing}, physiological signals \cite{ciftci2020fakecatcher} etc. As Datasets, they used lip-sync deep fakes that were created using Audio-to-Video \cite{suwajanakorn2017synthesizing} (A2V), Text-to-Video \cite{fried2019text} (T2V-S) and Text-to-Video (T2V-L) synthesis techniques. At first, they had to extract the target phonemes(MBP) using Google’s Speech-to-Text API \cite{iancu2019evaluating} in order to analyse a viseme during a spoken MBP phoneme. Then for a MBP occurrence in a video, total six frames were taken around the start of the occurrence. Using three approaches they tried to detect the veseme of a video. If the vesemes are completely closed while the MBP phonemes were spoken, then it is a real video otherwise a fake one. The first approach was totally manual, where an analyst manually labels all the sets of frames as “open” or “close” according to the mouths shape. This was used as the ground truth for the next two approaches. In the second one, they took the mouth region using 68 landmarks from a frame, converted to grayscale and measured the vertical intensity profile to predict if a given frame has mouth open or close. Finally, they used a Convolutional Neural Network (CNN) to classify if a mouth is open or closed in a single frame. Input of the network was a color image cropped around the mouth and rescaled to a 128 x 128 pixels, and the output was real values [0,1] corresponding to an “open” (0) or “closed” (1) mouth. Merging all three approaches gave them good accuracy on some datasets. For example, for the T2V-L dataset, 2nd approach gave 83.7\% and 3rd approach gave 71.1\% accuracy. Again for the T2V-S dataset, 2nd approach gave 89.5\% and 3rd approach gave 80.7\% accuracy.\\
In this research \cite{hsu2020deep}, the authors proposed a modified Convolutional network structure, including a pairwise learning approach, called the Common Fake Feature Network (CFFN) to detect fully generated fake images by Generative Adversarial Networks (GANs). As the other state-of-the-art fake image detectors are made of different variant of CNNs here they mainly focused on restructuring the networks with pairwise learning with an improved backbone network. In order to improve the performance of the proposed CFFN they used cross-layer feature investigation and to improve the generalization property the pairwise learning approach was introduced. The proposed two-step learning method combines the Common Fake Feature (CFF) based on pairwise learning strategy and the classifier learning uses two different loss functions for learning, 1) contrastive loss and 2) Binary cross-entropy loss. The contrastive loss learns the discriminative common fake feature (CFF) by the proposed CFFN and then the classification network captures the discriminative CFF to identify whether the image is real or fake. While constructing the Common Fake Feature Network(CFFN) Xception Network was used in \cite{marra2018detection} to capture the powerful feature from the training images in a purely supervised way. Besides, DenseNet and Siamese network architecture was used as a backbone network to achieve the discriminative CFF learning. In order to overcome the constant mapping problem of supervised learning models they changed the contrastive loss function a bit, and expressed as:\\
\normalsize
\begin{equation}
\scriptsize
L\left(W,\left(P, \mathbf{x}_{1}, \mathbf{x}_{2}\right)\right)=0.5 \times\left(y_{i j} E_{w}^{2}\right)+\left(1-y_{i j}\right) \times \max \left(0,\left(m-E_{w}\right)_{2}^{2}\right)
\end{equation}
\\
\normalsize
For the Classification learning they used a sub-network as a classifier so that, the cross-entropy loss function can be learned quickly. In a nutshell, the CFFN is first trained by the proposed contrastive loss and follows by training the classifier based on cross-entropy loss. Here the authors have used GAN generated images as initial datasets. Some of their used GAN models to create fake images were as follows.
\begin{enumerate}
\item WGAP (Wasserstein GAN)
\item WGAN-GP (WGAN with Gradient Penalty)
\item LSGAN (Least Squares GAN)  
\item PGGAN
\end{enumerate}
Finally, the gained accuracy for each of the datasets were:
\begin{enumerate}
\item 92.3\% for WGAP (Wasserstein GAN)
\item 92.7\% for WGAN-GP (WGAN with Gradient Penalty)
\item 94.7\% for LSGAN (Least Squares GAN)  
\item 98.3\% for PGGAN 
\end{enumerate}

This research \cite{mehra2020deepfake} is focused on detecting Deep Fake videos using VGG19 \cite{simonyan2014very} network as backbone with Capsule Network, using the capsule forensics model \cite{nguyen2019capsule} and LSTM Layers. They used capsule structure \cite{sabour2017dynamic} instead of CNN Layers as CNN layers max pool reduces features from images or frames of the video. To overcome these limitations Hinton et al. \cite{hinton2011transforming} in 2011 proposed the capsule architecture. Traditional CNN layers are not orientation independent where capsule networks are orientation independent. So, authors do not need to consider the frame of the rotated head or face as features of nose, mouth etc. are in the same orientation. Here authors used a capsule network to find inconsistency like blurred flickering of faces, lack of sharpness and reflection in different regions of the frame to detect spatial inconsistency in different frames of the videos to detect whether the video is fake or not. As Deep Fake algorithm creates fake videos frame by frame and it does not take consideration of the previous frame while creating a new frame for the fake videos so here this process creates some temporal inconsistency which the authors tried to detect to classify fake videos. The author applied the LSTM Network to detect inconsistency in multiple frames. They also used pre-processing techniques like cropping faces from the frames and searching specific faces in predefined bounding boxes to avoid scenarios like comparing one face with another rather than comparing same faces across multiple frames. They proposed pre-trained VGG19 model with Capsule Network with LSTM network to detect deep fake videos. They compared their approach with the state of the art XceptionNet Model \cite{rossler2019faceforensics++}. Their Capsule Network was not able to outperform the state of the art approach on DFDC \cite{dolhansky2020deepfake} dataset. Their performance or accuracy difference was about ~3.3\% ( without LSTM ) and ~1.7\% (both models with LSTM ). But Capsule Network+LSTM (4M parameters) was quite lighter than the state of the art XceptionNet+LSTM (27M parameters) about $\frac{1}{7}$th of XceptionNet+LSTM.

In this paper \cite{de2017detecting}, Authors Proposed a Transfer learning Approach to detect Computer. Generated Images. They proposed a model using ResNet-50 \cite{zhang2019detecting} by replacing the top 1000 fully connected softmax layers with 2 fully connected softmax layers. In this model pre-processed Computer Generated image passes in the training time to extract bottleneck features using their deep CNN layers. Average Pooling of this model generates the activation map of the bottleneck features. Then The bottleneck feature passes through the newly added softmax layer with freezing the Convolutional layers parameters to make this training much faster. They also proposed a second model where they just replace the top layer of ResNet-50 with an SVM \cite{bishop2006pattern} classifier rather than 2 fully connected softmax layers. The first proposed model (top 2 fully connected softmax layer) got average 0.941 Accuracy and the second model ( top SVM classifier ) got 92.3\% average accuracy on the dataset proposed by Tokuda et.al. \cite{tokuda2013computer}
In this paper \cite{jung2020deepvision} the author proposed an algorithm called DeepVision to detect GAN generated deepfakes by analyzing the eye blinking pattern. Deep Vision measures the repeated number of eye blinks in a specific time and verifies the anomaly based on that specific time period. The authors focused on find out the various factors which affect the eye blinking and based on that they recognised the anomaly. They asserted that the suggested algorithm operates only on the basis of pixels. DeepVision features a pre-processing phase in which it accepts input factors such as activity,gender, age and time to evaluate changes in human eye blink. Following that, it performs observations using the Target Detector and Eye Tracker. To monitor eye blinking, the system integrates Fast-HyperFace (facial detection) \cite{ranjan2017hyperface} with the EAR algorithm (eye detection) \cite{soukupova2016eye}. The Target Detector recognizes the item in the video, and the Eye Tracker records blinking using EAR (Eye-Aspect-Ratio)  \cite{soukupova2016eye}. The EAR computes the actual area of the horizontal and vertical axes using six points (pi) surrounding the eyes. The average of both eyes is then used to compute EAR. The ear can only perceive the blink of an eye smaller than the provided threshold value.

 The authors executed the procedures in a frame unit. They compared the measured datas with the actual human eye blink through DeepVision’s database and they called the method ‘Integrity Verification’. In the ‘Integrity Verification’ they compared the number of eye blinks, cycles, average cycle and the duration and after that they determined a video as fake if the resulting value is greater than the allowable range. However, the authors measured the benchmark using their dataset composed of deepfake videos and showed that seven out of eight deepfake videos were accurately identified which obtained an accuracy rate of 87.5\%. Lastly, they asserted that one of their limitations was that the integrity verification may not be applicable for the people having mental illnesses or nerve damage

In this paper \cite{li2018exposing}, a new deep learning algorithm is introduced to differentiate between fake videos created with Artificial Intelligent from real videos. The method author mentioned is centered on a case where the current deepfake algorithm is only able to generate certain predefined resolution images; thus, in order to properly swap or re-enact the target face, the source face is further warped to follow the original target face area. These modifications leave distinct artifacts in the produced deepfake footage, which CNN effectively detects. Unlike previous comparable efforts that combine vast numbers of actual and deepfake datasets to train the CNN, the author did not use deepfake generated datasets to train the CNN classifier as a negative training. The author's reasoning is that he was looking for artifacts in affine face warping to use as a distinguishing characteristic. Such artifacts can be easily mimicked by performing basic image processing operations on a picture to provide a negative example for training the CNN. As a result, this strategy takes less time and requires less resources to train. As positive examples, the CNN model's training phase is based on face images acquired from the internet (24,442 jpeg face images to be exact). As for negative examples instead of generating deepfake images the author used the proposed approach which simply generate the negative examples by affine face warping directly. The data is then pre-processed and then fed into the CNN models - VGG16 \cite{simonyan2014very}, RestNet50, RestNet101 and RestNet152. For each training batch, the author randomly selected half positive examples and converted them into negative examples following the aforementioned method of generating negative examples. The results were evaluated and validated on UADFV and DeepfakeTIMIT. The gained accuracy of the authors model in contrast to other models are listed below:
\begin{itemize}
\item VGG16 -  UADFV(84.5\%),  DeepFakeTIMIT(LQ- 84.6\%, HQ- 57.4\%)
\item RestNet50 - UADFV(97.4\%),  DeepFakeTIMIT(LQ- 99.9\%, HQ- 93.2\%)
\item RestNet101- UADFV(95.4\%),  DeepFakeTIMIT(LQ- 97.6\%, HQ- 86.9\%)
\item RestNet152- UADFV(93.8\%),  DeepFakeTIMIT(LQ-99.4\%, HQ- 91.2\%)
\end{itemize}

The author proposed a method built on the Expectation Maximization Algorithm in this paper \cite{guarnera2020fighting}. It was trained to extract and detect a fingerprint representing Convolutional traces (CT) left by a GAN during image generation. The CT has outstanding discriminative capabilities, surpassing the most recent models in the deepfake detection test while also demonstrating robustness in various approaches. The method was tested on various famous deepFake generation architecture such as CYCLEGAN, STARGAN, ATTGAN to name a few. The CT displays dependability and independence from image semantics, attaining a classification accuracy of more than 98\% while taking into account Deepfakes from 10 distinct GAN designs that are not just included in pictures of faces. Finally, testing on Deepfakes created by FACEAPP confirmed the usefulness of the suggested technique in a real-world setting, with 93\% in the fake detection job.

%% file: chapters/Work_Plan.tex
Deepfake video generation requires deep learning algorithms and techniques. In order to detect deepfake videos we will be using a deep neural network based approach to solve our deepfake video detection problem. We have planned to use traditional Convolutional Neural Networks with Capsule networks as More Convolutional Layers can cause significant data loss in video frames. So, we are trying to avoid this drawback by implementing CapsuleNetwork. We also want to use Long Short Term Memory (LSTM) technique because we are working with sequential video frames. The current methods that are available for deepfake video detection aim to solve the issue accurately but may fail to convince us of its reliability. Hence, we aim to use explainable Ai to clearly visualise the salient regions of the frames focused by the proposed model.\\
To summerize, the methodology proceeds by these following steps:\\
Step 1: Dataset Collection.\\
Step 2: Data Pre-processing.\\
Step 3: Data Labeling\\
Step 4: Data Splitting.\\
Step 5: Proposed CapsuleNet Model with LSTM.\\
Step 6: Creating Hybrid Model with pre-trained  models and CapsuleNet.\\
Step 7: Demonstrating performance of the proposed hybrid model.\\
Step 8: Comparison among the Proposed Model and the existing Models.\\

\begin{figure}[htbp]
\center
\includegraphics[width=0.5\textwidth]{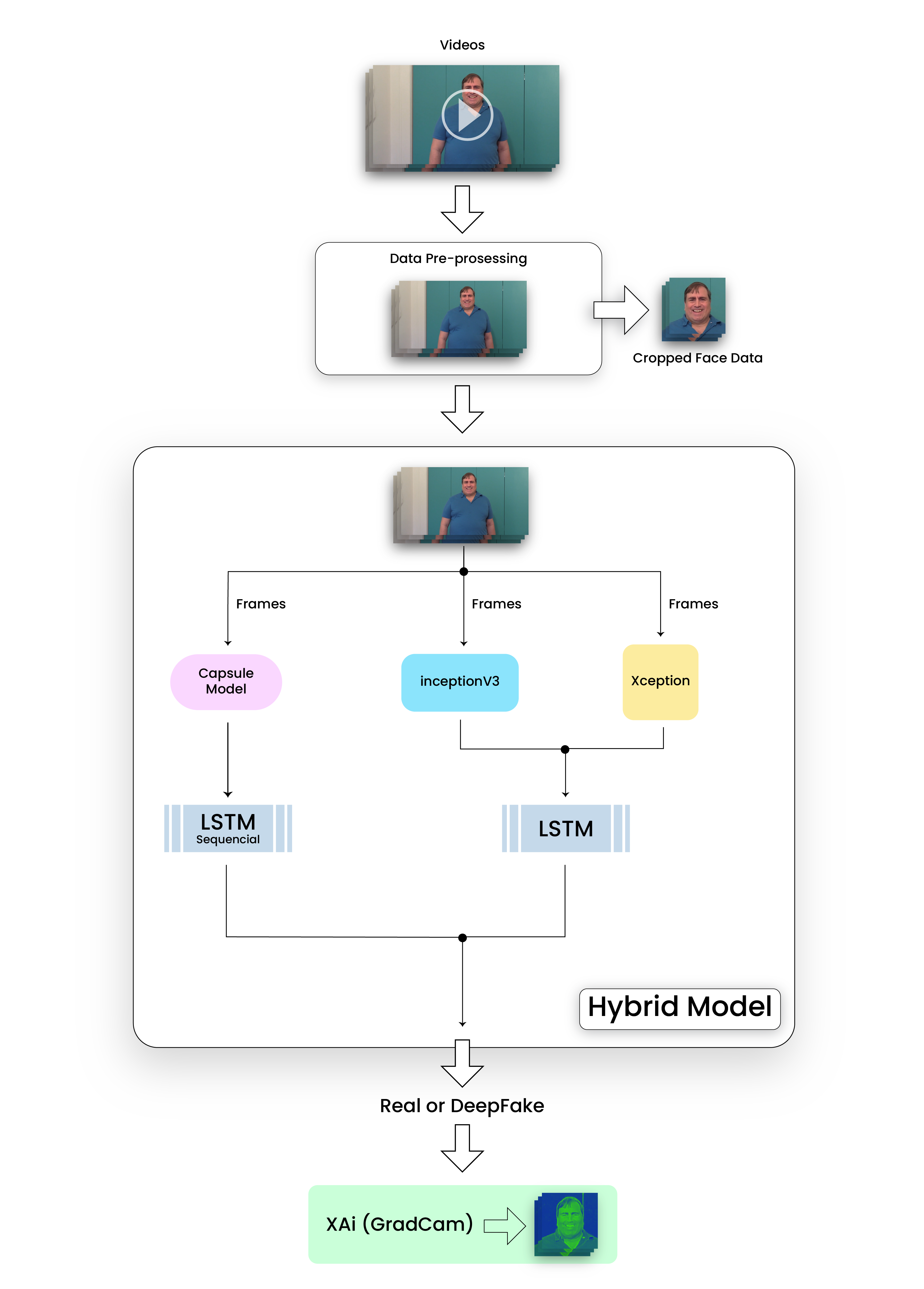}
\caption{Proposed Methodology Diagram}
\label{fig}
\end{figure}

%% file: chapters/Dataset.tex
We have chosen the DFDC dataset for our model. DFDC stands for Deepfake Detection Challenge \cite{dolhansky2020deepfake}. The DFDC dataset is considered the largest Deepfake detection dataset, which consists of more than 100,000 clips gathered from 3,426 paid actors. Each one of the 100,000 fake videos in the dataset is a unique target or source swap. The dataset provides publicly available face swap videos. The dataset is assembled using several Deepfake, GAN-based, and non-learned methods. The footages recorded in the DFDC dataset are primarily for the use of machine learning purposes. The dataset is the most substantial currently available dataset in terms of the number of videos and the number of frames. DFDC dataset consists of two versions. The first one is the Preview dataset which contains five thousand videos with two facial modification algorithm. Moreover, the second one is the Full dataset which contains one hundred and twenty-four thousand videos with eight facial modification algorithm. However, we worked with the Full dataset version.

\subsection{Data Pre-Processing}
 Data pre-processing is the preliminary step while designing a model. It diverts the raw data gathered from numerous sources into more relevant information. Data pre-processing is required because the raw data can consist of missing or inconsistent values and redundant information, which can inhibit the model from getting the desired outcome and cause substantial data loss. Moreover, most of the Deep Learning algorithms have a conventional input of target form. Before training, the dataset must be of that form.
The DFDC dataset consists of real and fake videos, with a .json file containing the labelling information. At first, We used the CV2 library to extract one frame per second from each video. We needed facial frames to train the model, so we used  DeepFace and DLIB libraries and cropped the facial frames. These libraries tend to lose resolution while cropping face frames from full frames, so we used both face and full frames for training. Two parts from the DFDC dataset, which approximately contains 5000 videos, were used to create our two datasets, one containing full-frame data another one containing only cropped face data. After preprocessing the data we chose ($5\times 128\times 128 \times 3$) as the target size, where frame=5, height = 128, width = 128 and channel = 3.


\begin{table}[h!]
\caption{Data Pre-processing}
\begin{tabularx}{0.49\textwidth}{|X|X|X|}
\hline \textbf{Data Segment} & \textbf{Parameters} & \textbf{Value} \\
\hline Training Data & Rescale & $1 . / 255$ \\
\cline { 2 - 3 } & Validation Split & $0.8$ \\
\cline { 2 - 3 } & Horizontal Flip & False \\
\cline { 2 - 3 } & Vertical flip & False \\
\hline { Validation Data } & Rescale & $1 . / 255$ \\
\cline { 2 - 3 } & Validation Split & $0.2$ \\
\hline Testing Data & Rescale & $1 . / 255$ \\
\hline
\end{tabularx}
\label{table:1}
\end{table}

\begin{figure}[htbp]
\center
{\includegraphics[scale=0.6]{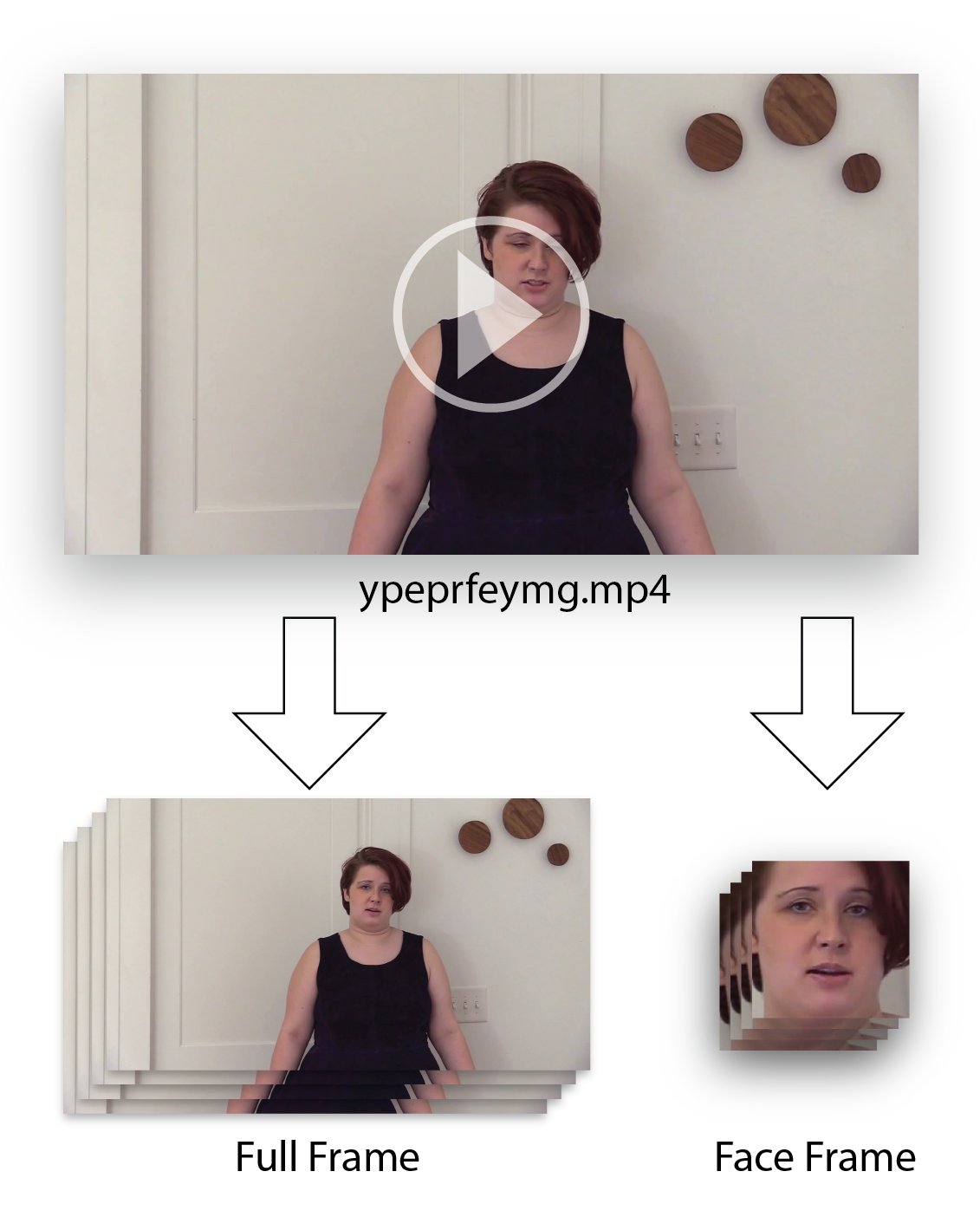}}
\caption{Data Pre-processing strategy}
\label{fig}
\end{figure} 

\subsection{Data Labeling}
The DFDC dataset was divided into more than 50 parts. Each of the parts was pre-labeled with a .json file inside it containing the labels of each video being real of fake. \\

\subsection{Data Splitting}
The data segmentation or splitting and the pre-processing were conducted simultaneously. The entire dataset was split into 80\% and 20\% for training data and testing data respectively based on an 8:2 ratio. Furthermore, 20\% of the training data were used for validation purposes. The split datasets have a batch size of 4, and the target size is ($5,128,128,3$). 4,948 videos have been used, and from each video, 5 frames on average were taken; therefore, approximately 24,740 frames have been used in the training, validation and testing. 

\begin{table}[h!]
\caption{Data Splitting}
\begin{tabularx}{0.49\textwidth}{|X|X|X|X|X|}
\hline
\textbf{Data Segments} &    \textbf{Percentage } & \textbf{Total Videos} & \textbf{Parameters} & \textbf{Value} \\ \hline
\multirow{4}{=}{Training} & \multirow{2}{=}{80\% of the total data} & \multirow{1}{=}{3958} & Target Size & (5,128,128,3) \\ \cline{4-5} 
                  &                   &               &   Class Mode   & Categorical \\ \cline{4-5} 
                  &                   &                   & Subset &  Training \\ \cline{4-5} 
                  &                   &                   & Batch Size & 4 \\ \hline
\multirow{4}{=}{Validation} & \multirow{2}{=}{20\% of Training data} & \multirow{1}{=}{792}  & Target Size & (5,128,128,3)  \\ \cline{4-5} 
                  &                   &                   & Class Mode & Categorical  \\ \cline{4-5} 
                  &                   &                   & Subset & Validation \\ \cline{4-5} 
                  &                   &                   & Batch Size & 4  \\ \hline
\multirow{2}{=}{Testing} & \multirow{1}{=}{20\% of the total data} & \multirow{1}{=}{989}  & Target Size & (5,128,128,3) \\ \cline{4-5} 
                  &                   &                   & Class Mode & Categorical  \\ \cline{4-5} 
                  &                   &                   & Subset & Testing \\ \cline{4-5} 
                  &                   &                   & Batch Size & 4  \\ \hline
\end{tabularx}

\label{table:1}
\end{table}


%% file: chapters/architecture.tex
\subsection{Proposed Hybrid model}
A Capsule Network consists of neurons which try to predict the location and the instantiation parameters of any particular entity using activity vectors. For example, an entity or entity portion \cite{sabour2017dynamic}. The probability of an object existing in a frame or image is denoted by the length parameter of the activity vector and the other parameters such as rotation, thickness are represented by the instantiation parameters. Any active capsule at one level anticipates the instantiation parameters of Capsules from its higher-level via routing by agreement method.
\begin{figure}[htbp]
\center
\includegraphics [width=0.5\textwidth] {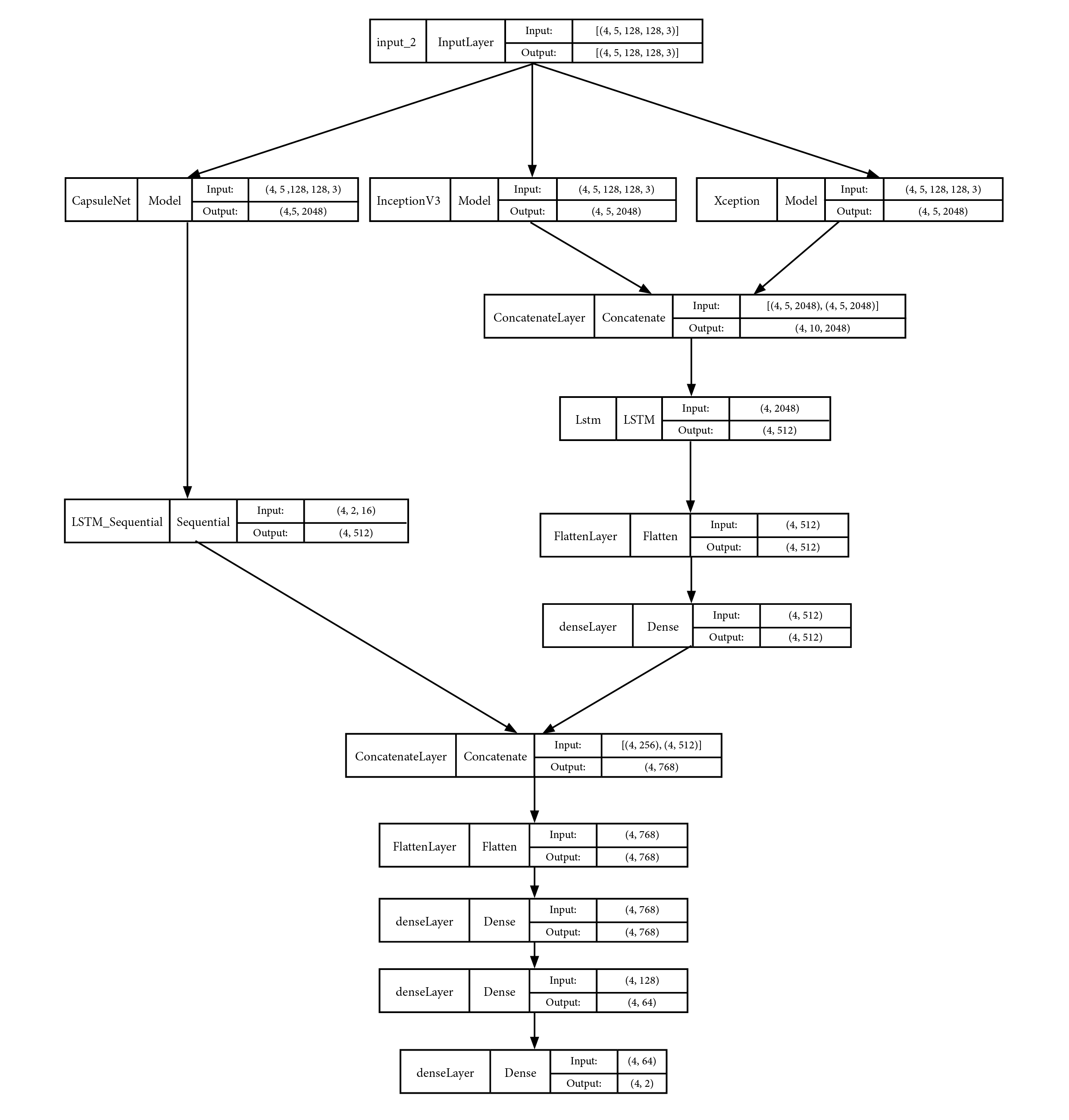}
\caption{Proposed Hybrid Model}
\label{fig}
\end{figure}

%% file: chapters/methodology.tex
\subsection{Convolutional LSTM\_2D Layer}
This is the first layer after the input layer. A ConvLSTM is similar to a LSTM layer, but the input transformations and recurrent transformations are both convolutional. As the next layer and traditionally the first layer of CapsuleNet is a 2D convolutional layer, this ConvLSTM layer maintains the spatiotemporal properties over time and gives input to the next convolutional 2D layer.

\subsection{Convolutional 2d Layer}
A traditional 2D convolutional layer, as proposed in the paper of Geoffrey E. Hinton \cite{sabour2017dynamic}. With a filter of 256, 9x9 convolution kernels, a stride of 1 and ReLU activation function, this layer translates pixel intensities into local feature detector activity, which are then fed into the primary capsule layer \cite{sabour2017dynamic}.

\subsection{Primary Capsule Layer}
Primary capsules can be seen as a convolutional layer with a Squash function to maintain the non-linearity. The convolutional layer outputs an array containing many feature maps, which are then reshaped to get \textit{M} vectors of \textit{N} dimensions representing features like rotations, translations, color, etc. In the last step, to make sure that no vector has a length longer than 1, we apply the Squash function \cite{sabour2017dynamic}. Since, the vector’s length represents the probability of an element being present in a particular position, so the summation of all the vectors should be 1. The Squash function given below makes sure of that.

$$
Squash function, \mathbf{v}_{j}=\frac{\left\|\mathbf{s}_{j}\right\|^{2}}{1+\left\|\mathbf{s}_{j}\right\|^{2}} \frac{\mathbf{s}_{j}}{\left\|\mathbf{s}_{j}\right\|}
$$


\subsection{Secondary Capsule Layer}
These capsules are our final classification capsules having 16 convolutional units. Each of the capsules of this layer is responsible for indicating the presence or absence of one particular class of the object such as the nose or the eyelashes or the lips. Every capsule in the first layer (primary caps) tries to predict the output of every capsule in the next layer (secondary caps) and thus send the weight matrix of that particular primary caps corresponding to that of the Secondary Caps. This process is called “routing-by-agreement” \cite{sabour2017dynamic} and this is where all the reshaping and squashing takes place in the whole algorithm. 

\subsection{Sequential LSTM Layer}
It is a standard LSTM layer that keeps track of the output of secondary capsule layers over time. LSTM’s are a special kind of RNN that work exceptionally well in terms of long-term dependencies. Here, the LSTM layer measures the consistency of the output classes representing different spatial component features. With the inconsistency of these features, we wish to determine if any manipulation had been done.

\subsection{Dense Layer}
The architecture consists of five dense layers, each of which passes on the outputs from the previous layers to the next neurons, with each neuron providing an output to the next layer. The last dense layer used as output layer with 2 neurons and softmax activation.
\pagebreak
\begin{table}[h!]
\caption{Summary of the proposed CapsuleNet model with LSTM.}
\begin{tabularx}{0.49\textwidth}{|X|X|X|X|}
\hline \multicolumn{2}{|l|}{\textbf{Layer(type)}} & \textbf{Output Shape} & \textbf{Parameter} \\
\hline \multicolumn{2}{|l|}{input\_1 (Input Layer)} & {$[(4,5,128,128,3)]$} & 0 \\
\hline \multicolumn{2}{|l|}{conv\_Ist\_m2d (ConvLSTM2D)} & $(4,128,128,128)$ & 604160 \\
\hline \multicolumn{2}{|l|}{conv1 (Conv2D)} & $(4,120,120,256)$ & 2654464 \\
\hline \multicolumn{2}{|l|}{primarycap\_conv2d (Conv2D)} & $(4,56,56,256)$ & 5308672 \\
\hline \multicolumn{2}{|l|}{primarycap\_reshape (Reshape)} & $(4,100352,8)$ & 0 \\
\hline \multicolumn{2}{|l|}{primarycap\_squash (Lambda)} & $(4,100352,8)$ & 0 \\
\hline \multicolumn{2}{|l|}{secondarycap (CapsuleLayer)} & $(4,2,16)$ & 25690112 \\
\hline \multicolumn{2}{|l|}{Istm\_1 (LSTM)} & (None, 1024) & 4263936 \\
\hline \multicolumn{2}{|l|}{dense\_1 (Dense)} & $($ None, 1024) & 1049600 \\
\hline \multicolumn{2}{|l|}{dense\_2 (Dense)} & (None, 512) & 524800 \\
\hline \multicolumn{2}{|l|}{dense\_3 (Dense)} & (None, 256) & 131328 \\
\hline \multicolumn{2}{|l|}{dense\_4 (Dense)} & (None, 64$)$ & 16448 \\
\hline \multicolumn{2}{|l|}{dense\_5 (Dense)(Output)} & (None, 2) & 130 \\
\hline
\end{tabularx}
\end{table}

\begin{table}[h!]
\begin{tabularx}{0.49\textwidth}{|X|X|X|}
\hline \textbf{Summary} & \textbf{Value} \\
\hline Total params & $40,243,650$ \\
\hline Trainable params & $40,243,650$ \\
\hline Non-trainable params & 0 \\
\hline
\end{tabularx}

\label{table:1}
\end{table}

\begin{table}[h!]
\caption{Summary of the pre-trained Xception model.}
\begin{tabularx}{0.49\textwidth}{|X|X|X|X|}
\hline \multicolumn{2}{|l|}{\textbf{Layer(type)}} & \textbf{Output Shape} & \textbf{Parameter} \\
\hline \multicolumn{2}{|l|}{input (InputLayer)} & {$[(None,128,128,3)]$} & 0 \\
\hline \multicolumn{2}{|l|}{block14\_sepconv2\_act} & $(None,8,8,2048)$ & 0 \\
\hline
\end{tabularx}
\end{table}
\begin{table}[h!]
\begin{tabularx}{0.49\textwidth}{|X|X|X|}
\hline \textbf{Summary} & \textbf{Value} \\
\hline Total params & $20, 861, 480$ \\
\hline Trainable params & $20, 806, 952$ \\
\hline Non-trainable params & $54, 528$ \\
\hline
\end{tabularx}
\label{table:1}
\end{table}

\begin{table}[h!]
\caption{Summary of the pre-trained InceptionV3 model.}
\begin{tabularx}{0.49\textwidth}{|X|X|X|X|}
\hline \multicolumn{2}{|l|}{\textbf{Layer(type)}} & \textbf{Output Shape} & \textbf{Parameter} \\
\hline \multicolumn{2}{|l|}{input (InputLayer)} & {$[(None,128,128,3)]$} & 0 \\
\hline \multicolumn{2}{|l|}{mixed10 (concatenate)} & $(None,6,6,2048)$ & 0 \\
\hline
\end{tabularx}
\end{table}
\begin{table}[h!]
\begin{tabularx}{0.49\textwidth}{|X|X|X|}
\hline \textbf{Summary} & \textbf{Value} \\
\hline Total params & $21, 802, 784$ \\
\hline Trainable params & $21, 768, 352$ \\
\hline Non-trainable params & $34, 432$ \\
\hline
\end{tabularx}
\label{table:1}
\end{table}

%% file: chapters/Implementation.tex
\subsection{Demonstrating performance of the model}
The implementation begins with training and evaluating the data epoch by epoch using hybrid model. While running the model, we took into account 30 epochs, batch size of 4, 4,948 videos and 5 frames each. The training took quite a time as we had to use Google Colab to perform the training operation. We evaluated the performance of our model using four parameters, Recall, Loss, Accuracy and AUC (Area Under ROC) (Receiver Operating Characteristics) curves.

\subsubsection{Result Analysis}
The following table \ref{table:performance} summarizes the performance of the implemented hybrid model:

\begin{table}[h!]
    \centering
        \caption{Performance of implemented Hybrid model}
    \begin{tabularx}{0.49\textwidth}{|X|X|}
       \hline \textbf{Parameter} & \textbf{Value (Full Frame)} \\
        \hline Accuracy (Validation) & $88.00\%$ \\
        \hline Loss (Validation) & $28.93\%$ \\
        \hline Recall (Validation) & $88.00\%$ \\
        \hline AUC (Validation) & $95.10\%$ \\
        \hline
    \end{tabularx}

    \label{table:performance}
\end{table}

\begin{figure}
    We have also plotted the performance in terms of training and validation to evaluate through a visual representation.\\
    
    \textbf{Accuracy:} These graphs represent the history of training and validation process accuracy. Our hybrid model performed well on the training set with an accuracy of 99.74\%. We can also see that our model gives about 88\% validation accuracy on the validation set. The validation accuracy indicates the actual accuracy of the model.\\
    \begin{center}
    {{\includegraphics[width=0.45\textwidth]{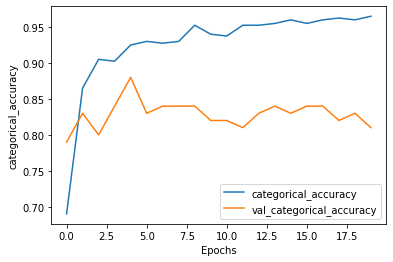} }}
    \end{center}
    \caption{History of Accuracy}
    \label{fig:accuracy}
\end{figure}

\begin{figure}
    \textbf{Loss:} The total error throughout all the 20 epochs are plotted in the graph. The loss graph reveals the error generated by the model and why the data could not be validated accross epochs which affects the precision and accuracy of the model, the validation loss is about 28.93\% . \\
    \begin{center}
    {{\includegraphics[width=0.45\textwidth]{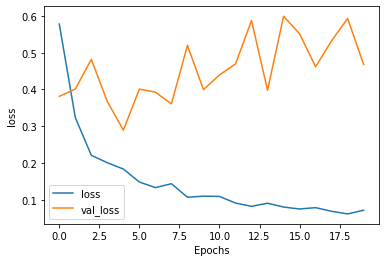} }}
    \end{center}
    \caption{History of Loss}
    \label{fig:loss}
\end{figure}

\begin{figure}
   \textbf{AUC:}  The Area Under the Curve (AUC) graph is a description of a model's Receiver Operating Characteristic (ROC) curve. It shows the capacity to discern the difference between positive and negative classifications. In the training data, our suggested model's AUC is about 99.74\% and in the validation data, it's around 95.10\%. This ensures that the AUC in this model is performing well.\\ 
   \begin{center}
    {{\includegraphics[width=0.45\textwidth]{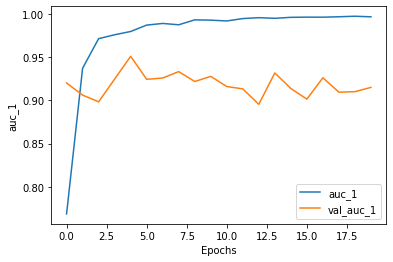} }}
   \end{center}
    
    \caption{History of AUC}
    \label{fig:auc}
\end{figure}

\begin{figure}
    \textbf{Recall:}  By definition, Recall answers the question that, if proportion of actual positives was identified correctly or not. It means the percentage of a certain class correctly identified from all of the given examples of that class. Training recall value of our model was 96.50\% and validation recall value was 88\%.
    \begin{center}
        {{\includegraphics[width=0.45\textwidth]{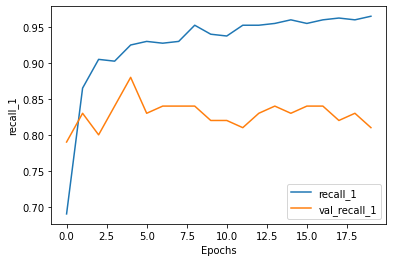} }}
    \end{center}
   
    \caption{History of Recall}
    \label{fig:recall}
\end{figure}


\begin{figure}
 \textbf{Explainable Ai:} An explainable Ai (XAi) system's goal is to make a model's behavior more understandable to humans by providing explanations. The purpose of Gradient-weighted Class Activation Mapping (Grad-CAM) is to make the convolutional neural networks more transparent by visualizing input regions which are relevant and important for predictions or for visual explanations. It exploits the spatial information collected by the convolutional layers to identify which regions of a source image were significant for a classification decision. In the heatmap the red color resembles the regions where activation weight is high and the blue regions are where activation weight is low. From our model we can have the explanation as such, In the real videos the activation regions are found in the facial areas reflecting that the model has depicted proficient facial features to classify it as a real video. In contrast, activation regions are not seen around the facial areas in fake videos depicting that the model did not find promising facial features to classify it as a real video. Therefore, it classifies the frame as fake. Additionally,
   The pre-trained models are not specifically targeted to work only on faces. Therefore, by using XAi, we wanted to make sure that our model is focusing on the faces as we have combined the pre-trained models with our model.

   \begin{center}
    {{\includegraphics[width=0.5\textwidth]{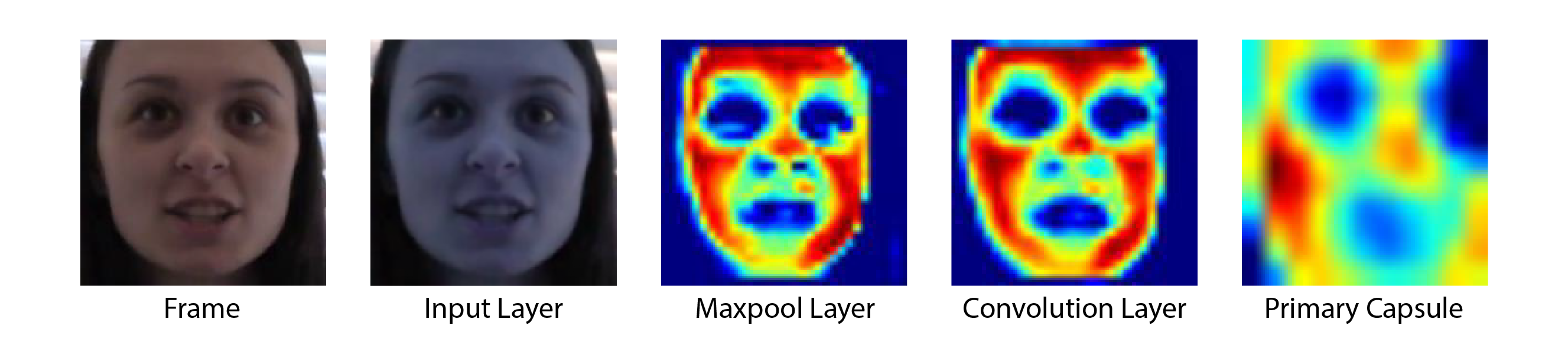} }}
   \end{center}
    \caption{Grad-CAM Activation heatmap (Real video)}
    \label{fig:gradReal}
    
     \begin{center}
    {{\includegraphics[width=0.5\textwidth]{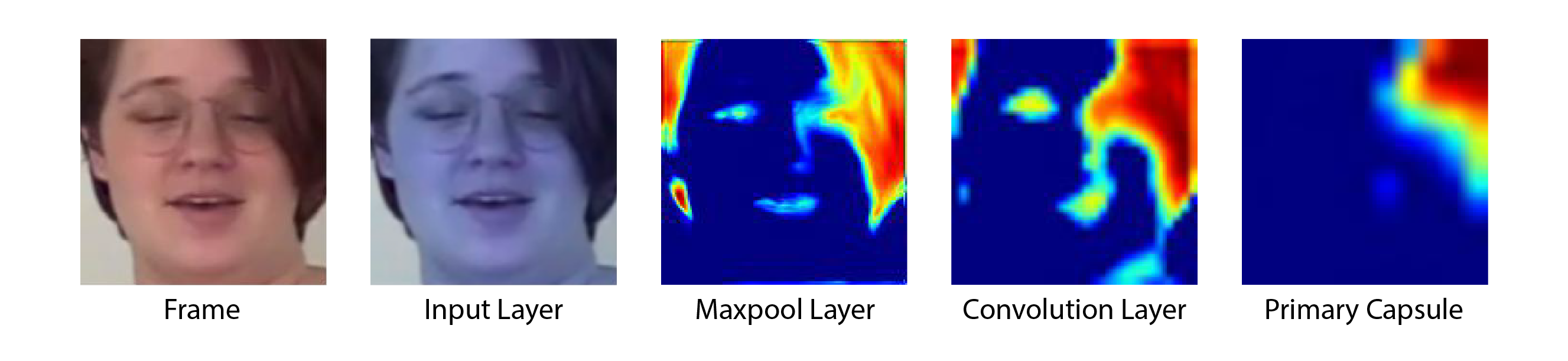} }}
   \end{center}
    \caption{Grad-CAM Activation heatmap (Fake video)}
    \label{fig:gradFake}
    
\end{figure}

%% file: chapters/Analysis.tex
To compare the dynamics of our hybrid model we want to highlight a model which uses the same dataset as ours and follows a combined model approach  similar to ours.\\

On the other hand, to measure the performance of the classification process they followed a combined model approach using three detection models. They used TemperNet, a frame-based model as the first model. TamperNet is trained to detect low level image manipulations compiled with a small DNN which was made of 6 convolutional layers and a 1 fully connected layer, for example placing objects or adding digital text to an image. XceptionNet face detection and full-image models are the others, which were trained using the FaceForensics data sets. Choosing a sample of one frame per second of videos, this approach achieved an accuracy rate of 83.3\% on TemperNet, 93\% and 78.4\% on face and full frame using XceptionNet respectively. \\

\begin{table}[h!]
    \caption{Comparison among the proposed model and combined model}
\begin{tabularx}{0.49\textwidth}{|X|X|X|X|X|}
\hline \multicolumn{3}{|l|}{\textbf{Model}} & \textbf{Accuracy}  & \textbf{Recall}  \\
\hline \multicolumn{3}{|l|}{Combined Model} & $83.30 \%$  & $3.30 \%$ \\
\hline \multicolumn{3}{|l|}{Proposed Model} & $88.00 \%$ & $88.30 \%$  \\
\hline
\end{tabularx}
\end{table}

Our hybrid model has shown significantly improved results on the DFDC dataset compared to the above mentioned combined model approach. While considering full frame input sets our model shows nearly 5\% more accuracy.

%% file: chapters/Conclusion.tex
\section{Limitations \& Future Works}
Our model has shown promising results with the given dataset. However, we are looking forward to using different datasets to feed into our model to see how it performs. Moreover, we want to merge multiple datasets and train our model to ensure unbiased deep learning. In the future, we would like to improve the accuracy by changing the frame's properties such as brightness, contrast, and gray scale. Additionally, we would like to work on different frame rates as well. We can increase our current frame rate which will enlarge our dataset and can be beneficial to improve our model’s accuracy. We would also like to work on other meta areas that we have reviewed in various papers such as, audio spoofing, tracking iris and nose pointer locations. As we have implemented explainable Ai, we would like to work on the activation heatmaps and try to find a pattern of our model of detecting fake videos. We look forward to working on these shortcomings and scopes to research further to improve our model.

\section{Conclusion}
With the rise of Deepfake videos, this study provides a comprehensive overview of recent development in fake media generation and their danger to online information's authenticity and trustworthiness. We describe the deepfake generation methods, the inconsistencies introduced by the AI, available data-sets and how current deepfake detection techniques exploit these inconsistencies. Here we introduce a new hybrid model using traditional Convolutional Neural Networks and Capsule Networks integrated with LSTM. We used pre-processed frames of the video focusing especially on the facial region to train our model as most of the changes occur in the facial region for the deepfake videos.
For both real and fake videos CNN based Capsule Network was used to detect feature vectors and LSTM layer used these feature vectors including spatial inconsistencies to identify temporal discrepancies within these frames and forecast whether the video from which these features were collected is real or fake. In future, we will try to answer the question “why” are the videos which will be detected as fake are actually fake. In other words, what are the features that makes these videos fake by extending our model with the Explainable AI.

%% file: chapters/Ackowledgement.tex
All praised to our Almighty Allah for whom we completed our work without any major intervention. Then we would like to thank our supervisor Dr. Mohammad Zavid Parvez Sir and co-supervisor Md. Tanzim Reza Sir for their assistance and advice. Without their continuous support, we would not be able to make it to the end. Lastly, our praiseworthy team coordination has helped us to reach the finish line.

%% file: references.bib
@article{holmes2016assessing,
  title={Assessing and improving the identification of computer-generated portraits},
  author={Holmes, Olivia and Banks, Martin S and Farid, Hany},
  journal={ACM Transactions on Applied Perception (TAP)},
  volume={13},
  number={2},
  pages={1--12},
  year={2016},
  publisher={ACM New York, NY, USA}
}

@inproceedings{korshunova2017fast,
  title={Fast face-swap using convolutional neural networks},
  author={Korshunova, Iryna and Shi, Wenzhe and Dambre, Joni and Theis, Lucas},
  booktitle={Proceedings of the IEEE international conference on computer vision},
  pages={3677--3685},
  year={2017}
}

@inproceedings{thies2016face2face,
  title={Face2face: Real-time face capture and reenactment of rgb videos},
  author={Thies, Justus and Zollhofer, Michael and Stamminger, Marc and Theobalt, Christian and Nie{\ss}ner, Matthias},
  booktitle={Proceedings of the IEEE conference on computer vision and pattern recognition},
  pages={2387--2395},
  year={2016}
}

@article{lorenzo2018voice,
  title={The voice conversion challenge 2018: Promoting development of parallel and nonparallel methods},
  author={Lorenzo-Trueba, Jaime and Yamagishi, Junichi and Toda, Tomoki and Saito, Daisuke and Villavicencio, Fernando and Kinnunen, Tomi and Ling, Zhenhua},
  journal={arXiv preprint arXiv:1804.04262},
  year={2018}
}

@inproceedings{afchar2018mesonet,
  title={Mesonet: a compact facial video forgery detection network},
  author={Afchar, Darius and Nozick, Vincent and Yamagishi, Junichi and Echizen, Isao},
  booktitle={2018 IEEE International Workshop on Information Forensics and Security (WIFS)},
  pages={1--7},
  year={2018},
  organization={IEEE}
}

@inproceedings{szegedy2017inception,
  title={Inception-v4, inception-resnet and the impact of residual connections on learning},
  author={Szegedy, Christian and Ioffe, Sergey and Vanhoucke, Vincent and Alemi, Alexander},
  booktitle={Proceedings of the AAAI Conference on Artificial Intelligence},
  volume={31},
  number={1},
  year={2017}
}

@inproceedings{guera2018deepfake,
  title={Deepfake video detection using recurrent neural networks},
  author={G{\"u}era, David and Delp, Edward J},
  booktitle={2018 15th IEEE International Conference on Advanced Video and Signal Based Surveillance (AVSS)},
  pages={1--6},
  year={2018},
  organization={IEEE}
}

@article{goodfellow2014generative,
  title={Generative adversarial nets. In NIPS},
  author={Goodfellow, I and Pouget-Abadie, J and Mirza, M and Xu, B and Warde-Farley, D and Ozair, S and Courville, A and Bengio, Y},
  year={2014}
}

@article{denton2015deep,
  title={Deep generative image models using a laplacian pyramid of adversarial networks},
  author={Denton, Emily and Chintala, Soumith and Szlam, Arthur and Fergus, Rob},
  journal={arXiv preprint arXiv:1506.05751},
  year={2015}
}

@inproceedings{shrivastava2017learning,
  title={Learning from simulated and unsupervised images through adversarial training},
  author={Shrivastava, Ashish and Pfister, Tomas and Tuzel, Oncel and Susskind, Joshua and Wang, Wenda and Webb, Russell},
  booktitle={Proceedings of the IEEE conference on computer vision and pattern recognition},
  pages={2107--2116},
  year={2017}
}

@article{liu2017unsupervised,
  title={Unsupervised image-to-image translation networks},
  author={Liu, Ming-Yu and Breuel, Thomas and Kautz, Jan},
  journal={arXiv preprint arXiv:1703.00848},
  year={2017}
}

@inproceedings{zhu2017unpaired,
  title={Unpaired image-to-image translation using cycle-consistent adversarial networks},
  author={Zhu, Jun-Yan and Park, Taesung and Isola, Phillip and Efros, Alexei A},
  booktitle={Proceedings of the IEEE international conference on computer vision},
  pages={2223--2232},
  year={2017}
}

@inproceedings{bansal2018recycle,
  title={Recycle-gan: Unsupervised video retargeting},
  author={Bansal, Aayush and Ma, Shugao and Ramanan, Deva and Sheikh, Yaser},
  booktitle={Proceedings of the European conference on computer vision (ECCV)},
  pages={119--135},
  year={2018}
}

@inproceedings{choi2018stargan,
  title={Stargan: Unified generative adversarial networks for multi-domain image-to-image translation},
  author={Choi, Yunjey and Choi, Minje and Kim, Munyoung and Ha, Jung-Woo and Kim, Sunghun and Choo, Jaegul},
  booktitle={Proceedings of the IEEE conference on computer vision and pattern recognition},
  pages={8789--8797},
  year={2018}
}

@inproceedings{chollet2017xception,
  title={Xception: Deep learning with depthwise separable convolutions},
  author={Chollet, Fran{\c{c}}ois},
  booktitle={Proceedings of the IEEE conference on computer vision and pattern recognition},
  pages={1251--1258},
  year={2017}
}

@inproceedings{he2016deep,
  title={Deep residual learning for image recognition},
  author={He, Kaiming and Zhang, Xiangyu and Ren, Shaoqing and Sun, Jian},
  booktitle={Proceedings of the IEEE conference on computer vision and pattern recognition},
  pages={770--778},
  year={2016}
}

@article{sabir2019recurrent,
  title={Recurrent convolutional strategies for face manipulation detection in videos},
  author={Sabir, Ekraam and Cheng, Jiaxin and Jaiswal, Ayush and AbdAlmageed, Wael and Masi, Iacopo and Natarajan, Prem},
  journal={Interfaces (GUI)},
  volume={3},
  number={1},
  year={2019}
}

@inproceedings{prasad2006resampling,
  title={On resampling detection and its application to detect image tampering},
  author={Prasad, Santasriya and Ramakrishnan, KR},
  booktitle={2006 IEEE International Conference on Multimedia and Expo},
  pages={1325--1328},
  year={2006},
  organization={IEEE}
}

@article{kirchner2008hiding,
  title={Hiding traces of resampling in digital images},
  author={Kirchner, Matthias and Bohme, Rainer},
  journal={IEEE Transactions on Information Forensics and Security},
  volume={3},
  number={4},
  pages={582--592},
  year={2008},
  publisher={IEEE}
}

@article{wang2019fakespotter,
  title={FakeSpotter: A simple yet robust baseline for spotting AI-synthesized fake faces},
  author={Wang, Run and Juefei-Xu, Felix and Ma, Lei and Xie, Xiaofei and Huang, Yihao and Wang, Jian and Liu, Yang},
  journal={arXiv preprint arXiv:1909.06122},
  year={2019}
}

@inproceedings{zhang2019detecting,
  title={Detecting and simulating artifacts in gan fake images},
  author={Zhang, Xu and Karaman, Svebor and Chang, Shih-Fu},
  booktitle={2019 IEEE International Workshop on Information Forensics and Security (WIFS)},
  pages={1--6},
  year={2019},
  organization={IEEE}
}

@article{radford2015unsupervised,
  title={Unsupervised representation learning with deep convolutional generative adversarial networks},
  author={Radford, Alec and Metz, Luke and Chintala, Soumith},
  journal={arXiv preprint arXiv:1511.06434},
  year={2015}
}

@article{moon1996expectation,
  title={The expectation-maximization algorithm},
  author={Moon, Todd K},
  journal={IEEE Signal processing magazine},
  volume={13},
  number={6},
  pages={47--60},
  year={1996},
  publisher={IEEE}
}

@inproceedings{agarwal2020detecting,
  title={Detecting deep-fake videos from phoneme-viseme mismatches},
  author={Agarwal, Shruti and Farid, Hany and Fried, Ohad and Agrawala, Maneesh},
  booktitle={Proceedings of the IEEE/CVF Conference on Computer Vision and Pattern Recognition Workshops},
  pages={660--661},
  year={2020}
}

@inproceedings{li2018ictu,
  title={In ictu oculi: Exposing ai created fake videos by detecting eye blinking},
  author={Li, Yuezun and Chang, Ming-Ching and Lyu, Siwei},
  booktitle={2018 IEEE International Workshop on Information Forensics and Security (WIFS)},
  pages={1--7},
  year={2018},
  organization={IEEE}
}

@inproceedings{yang2019exposing,
  title={Exposing deep fakes using inconsistent head poses},
  author={Yang, Xin and Li, Yuezun and Lyu, Siwei},
  booktitle={ICASSP 2019-2019 IEEE International Conference on Acoustics, Speech and Signal Processing (ICASSP)},
  pages={8261--8265},
  year={2019},
  organization={IEEE}
}

@article{ciftci2020fakecatcher,
  title={Fakecatcher: Detection of synthetic portrait videos using biological signals},
  author={Ciftci, Umur Aybars and Demir, Ilke and Yin, Lijun},
  journal={IEEE Transactions on Pattern Analysis and Machine Intelligence},
  year={2020},
  publisher={IEEE}
}

@article{suwajanakorn2017synthesizing,
  title={Synthesizing obama: learning lip sync from audio},
  author={Suwajanakorn, Supasorn and Seitz, Steven M and Kemelmacher-Shlizerman, Ira},
  journal={ACM Transactions on Graphics (ToG)},
  volume={36},
  number={4},
  pages={1--13},
  year={2017},
  publisher={ACM New York, NY, USA}
}

@article{fried2019text,
  title={Text-based editing of talking-head video},
  author={Fried, Ohad and Tewari, Ayush and Zollh{\"o}fer, Michael and Finkelstein, Adam and Shechtman, Eli and Goldman, Dan B and Genova, Kyle and Jin, Zeyu and Theobalt, Christian and Agrawala, Maneesh},
  journal={ACM Transactions on Graphics (TOG)},
  volume={38},
  number={4},
  pages={1--14},
  year={2019},
  publisher={ACM New York, NY, USA}
}

@article{iancu2019evaluating,
  title={Evaluating Google Speech-to-Text API's Performance for Romanian e-Learning Resources.},
  author={Iancu, Bogdan},
  journal={Informatica Economica},
  volume={23},
  number={1},
  year={2019}
}

@article{hsu2020deep,
  title={Deep fake image detection based on pairwise learning},
  author={Hsu, Chih-Chung and Zhuang, Yi-Xiu and Lee, Chia-Yen},
  journal={Applied Sciences},
  volume={10},
  number={1},
  pages={370},
  year={2020},
  publisher={Multidisciplinary Digital Publishing Institute}
}

@inproceedings{marra2018detection,
  title={Detection of gan-generated fake images over social networks},
  author={Marra, Francesco and Gragnaniello, Diego and Cozzolino, Davide and Verdoliva, Luisa},
  booktitle={2018 IEEE Conference on Multimedia Information Processing and Retrieval (MIPR)},
  pages={384--389},
  year={2018},
  organization={IEEE}
}

@mastersthesis{mehra2020deepfake,
  title={Deepfake detection using capsule networks with long short-term memory networks},
  author={Mehra, Akul},
  year={2020},
  school={University of Twente}
}

@article{simonyan2014very,
  title={Very deep convolutional networks for large-scale image recognition},
  author={Simonyan, Karen and Zisserman, Andrew},
  journal={arXiv preprint arXiv:1409.1556},
  year={2014}
}

@inproceedings{nguyen2019capsule,
  title={Capsule-forensics: Using capsule networks to detect forged images and videos},
  author={Nguyen, Huy H and Yamagishi, Junichi and Echizen, Isao},
  booktitle={ICASSP 2019-2019 IEEE International Conference on Acoustics, Speech and Signal Processing (ICASSP)},
  pages={2307--2311},
  year={2019},
  organization={IEEE}
}

@article{sabour2017dynamic,
  title={Dynamic routing between capsules},
  author={Sabour, Sara and Frosst, Nicholas and Hinton, Geoffrey E},
  journal={arXiv preprint arXiv:1710.09829},
  year={2017}
}

@inproceedings{hinton2011transforming,
  title={Transforming auto-encoders},
  author={Hinton, Geoffrey E and Krizhevsky, Alex and Wang, Sida D},
  booktitle={International conference on artificial neural networks},
  pages={44--51},
  year={2011},
  organization={Springer}
}

@inproceedings{rossler2019faceforensics++,
  title={Faceforensics++: Learning to detect manipulated facial images},
  author={Rossler, Andreas and Cozzolino, Davide and Verdoliva, Luisa and Riess, Christian and Thies, Justus and Nie{\ss}ner, Matthias},
  booktitle={Proceedings of the IEEE/CVF International Conference on Computer Vision},
  pages={1--11},
  year={2019}
}

@article{dolhansky2020deepfake,
  title={The deepfake detection challenge dataset},
  author={Dolhansky, Brian and Bitton, Joanna and Pflaum, Ben and Lu, Jikuo and Howes, Russ and Wang, Menglin and Ferrer, Cristian Canton},
  journal={arXiv preprint arXiv:2006.07397},
  year={2020}
}

@inproceedings{de2017detecting,
  title={Detecting computer generated images with deep convolutional neural networks},
  author={De Rezende, Edmar RS and Ruppert, Guilherme CS and Carvalho, Tiago},
  booktitle={2017 30th SIBGRAPI Conference on Graphics, Patterns and Images (SIBGRAPI)},
  pages={71--78},
  year={2017},
  organization={IEEE}
}

@book{bishop2006pattern,
  title={Pattern recognition and machine learning},
  author={Bishop, Christopher M},
  year={2006},
  publisher={springer}
}

@article{tokuda2013computer,
  title={Computer generated images vs. digital photographs: A synergetic feature and classifier combination approach},
  author={Tokuda, Eric and Pedrini, Helio and Rocha, Anderson},
  journal={Journal of Visual Communication and Image Representation},
  volume={24},
  number={8},
  pages={1276--1292},
  year={2013},
  publisher={Elsevier}
}

@article{jung2020deepvision,
  title={DeepVision: Deepfakes Detection Using Human Eye Blinking Pattern},
  author={Jung, Tackhyun and Kim, Sangwon and Kim, Keecheon},
  journal={IEEE Access},
  volume={8},
  pages={83144--83154},
  year={2020},
  publisher={IEEE}
}

@article{ranjan2017hyperface,
  title={Hyperface: A deep multi-task learning framework for face detection, landmark localization, pose estimation, and gender recognition},
  author={Ranjan, Rajeev and Patel, Vishal M and Chellappa, Rama},
  journal={IEEE transactions on pattern analysis and machine intelligence},
  volume={41},
  number={1},
  pages={121--135},
  year={2017},
  publisher={IEEE}
}

@inproceedings{soukupova2016eye,
  title={Eye blink detection using facial landmarks},
  author={Soukupova, Tereza and Cech, Jan},
  booktitle={21st computer vision winter workshop, Rimske Toplice, Slovenia},
  year={2016}
}

@article{li2018exposing,
  title={Exposing deepfake videos by detecting face warping artifacts},
  author={Li, Yuezun and Lyu, Siwei},
  journal={arXiv preprint arXiv:1811.00656},
  year={2018}
}

@article{guarnera2020fighting,
  title={Fighting Deepfake by Exposing the Convolutional Traces on Images},
  author={Guarnera, Luca and Giudice, Oliver and Battiato, Sebastiano},
  journal={IEEE Access},
  volume={8},
  pages={165085--165098},
  year={2020},
  publisher={IEEE}
}

@Misc{obama,
  author =   {BuzzFeedVideo},
  title =    {You Won’t Believe What Obama Says In This Video!},
  howpublished = {Available: \url{https://www.youtube.com/watch?v=cQ54GDm1eL0}},
  month =    {April},
  day =          17,
  year =     2018
}

@article{ galindo_2020, title={XR Belgium posts deepfake of Belgian premier linking Covid-19 with climate crisis}, url={https://www.brusselstimes.com/news/belgium-all-news/politics/106320/xr-belgium-posts-deepfake-of-belgian-premier-linking-covid-19-with-climate-crisis/}, journal={The Brussels Times}, author={ Galindo, Gabriela}, year={2020}, month={April}}
